\def\year{2019}\relax
\documentclass[letterpaper]{article} %DO NOT CHANGE THIS
\usepackage{aaai19}  %Required
\usepackage{times}  %Required
\usepackage{helvet}  %Required
\usepackage{courier}  %Required
\usepackage{url}  %Required
\usepackage{graphicx}  %Required

\usepackage{multirow}
\usepackage{epsfig}
\usepackage{amsmath}
\usepackage{amssymb,amsfonts}
\usepackage{bm}
\usepackage{CJK}
\usepackage{subcaption}
\usepackage{float}
\usepackage{adjustbox}
\usepackage{makecell}
\usepackage{dsfont}

\newcommand{\citet}[1]{\citeauthor{#1}~\shortcite{#1}}
\newcommand{\citep}{\cite}

\nocopyright

\begin{document}

\title{Implanting Rational Knowledge into Distributed Representation \\at Morpheme Level}
\author{Zi Lin\textsuperscript{1,3} \and
  Yang Liu\textsuperscript{2,3} \\
  \textsuperscript{1}Department of Chinese Language and Literature, Peking University\\
  \textsuperscript{2}Institute of Computational Linguistics, Peking University\\
  \textsuperscript{3}Key Laboratory of Computational Linguistics (Ministry of Education), Peking University\\
  \{zi.lin, liuyang\}@pku.edu.cn \\}

\maketitle
\begin{abstract}
Previously, researchers paid no attention to the creation of unambiguous morpheme embeddings independent from the corpus, while such information plays an important role in expressing the exact meanings of words for parataxis languages like Chinese. In this paper, after constructing the Chinese lexical and semantic ontology based on word-formation, we propose a novel approach to implanting the structured rational knowledge into distributed representation at morpheme level, naturally avoiding heavy disambiguation in the corpus. We design a template to create the instances as pseudo-sentences merely from the pieces of knowledge of morphemes built in the lexicon. To exploit hierarchical information and tackle the data sparseness problem, the instance proliferation technique is applied based on similarity to expand the collection of pseudo-sentences. The distributed representation for morphemes can then be trained on these pseudo-sentences using word2vec. For evaluation, we validate the paradigmatic and syntagmatic relations of morpheme embeddings, and apply the obtained embeddings to word similarity measurement, achieving significant improvements over the classical models by more than 5 Spearman scores or 8 percentage points, which shows very promising prospects for adoption of the new source of knowledge.
\end{abstract}
\begin{CJK}{UTF8}{gbsn}
\section{Introduction}
Nowadays, learning representations for the meanings of words has been a key problem in natural language processing (NLP).  The basic unit in NLP is usually and mainly word. However, for parataxis languages like Chinese, which is made up of hieroglyphic characters, {\it word} is not a natural unit, and character can provide yet rich semantic information \cite{Fu-81,Xu-04}.

Theoretically, the meanings of Chinese words can be deduced from the meanings of characters. However, the same Chinese characters within words may hold different meanings, so the meanings of an identical character should be further differentiated. For example, in the words ``花钱'' (to spend money) and ``桃花'' (peach-blossom), the character meanings of ``花'' are not the same. Chinese linguists use the term \textit{yusu} (morpheme) to distinguish identical characters with different meanings, which is defined as \textit{the smallest combination of meaning and phonetic sound in Chinese} \cite{ZD-82}. Previously, linguists define \textit{morpheme} as \textit{the smallest meaning-bearing unit of language as well as the smallest unit of syntax} \cite{matthews1972inflectional}. The Chinese morpheme is close to this definition in terms of semantics but requires a smallest phonetic sound, i.e., it should correspond to a character in form. Therefore, the character ``花'' in ``花钱'' and ``桃花'' refers to different morphemes ``花$_1$'' and ``花$_2$''. We sometimes use the term sememe to refer to the meanings of morphemes, so the sememe of ``花$_1$'' is \textit{to spend}, while the sememe of ``花$_2$'' is \textit{flowers}.

The meanings of words and the meanings of morphemes are highly related to some extent, and the sub-units of words will form patterns during word-building. For example, in Chinese words ``桃花'' (peach-blossom), ``樱花'' (cherry-blossom) and ``荷花'' (lotus-blossom), one will find the character ``花'', which means \textit{flowers}, as a common component. The components before ``花'', as different modifiers, also hold their meanings. These words are of Modifier-Head structure, and the morphology is somewhat similar to that of syntax in Chinese. 

Therefore, morphemes and their combination patterns are very important for the generation of the meanings of words. Researchers have addressed the importance of morphological compositionality and word-formation analysis. \cite{lazaridou2013compositional} explored the application of compositional distributed semantic models, originally designed to learn phrase meanings. \cite{luong2013better} combined recursive neural networks with neural language models to consider contextual information in learning morphologically-aware word representations.

As for Chinese, \cite{chen2015joint} proposed a character-enhanced word embeddings model (CWE) by regarding the word as the simple combination of characters and obtained character embeddings on the corpus, without further knowledge of morphemes or sememes input. \cite{niu2017improved} employed HowNet to learn word representation on corpus for the SE-WRL model. By using attention schema for rough word sense disambiguation, HowNet sememe embeddings were thus obtained directly from the large-scale corpus. 

To avoid sense disambiguation on the corpus, there was also work trying to obtain word sense embeddings by leveraging the dictionary definitions or lexical resources. {\tt Dict2vec} used the English version of Cambridge, Oxford, Collins and dictionary.com to build new word pairs so that semantically-related words are moved closer, and negative sampling filters out pairs whose words are unrelated in dictionaries \cite{tissier2017dict2vec}. As there is no structured knowledge can be exploited in the dictionaries, such selection of word pairs tended to be cumbersome and complicated. {\tt WordNet2vec} \cite{bartusiak2017wordnet2vec} created vectors for each word from WordNet by first simplifying the WordNet structure into a graph and then utilizing the shortest paths to encode the words, which cannot distinguish and retain the specific semantic relations.

To the best of our knowledge, all these works aimed to get better representations for words, while little emphasis has been laid on creation and analysis of relatively independent morpheme embeddings. Also, all the works relied highly on corpora or simple dictionary knowledge, and have not been able to form distributed representation directly from the structured rational knowledge.

In this paper, after constructing the Chinese lexical and semantic ontology based on word-formation, we propose a novel approach to implanting the structured rational knowledge into distributed representation at morpheme level without using any text corpus. We first introduce the construction of rational knowledge of Chinese morphemes at Peking University. Then we extract this knowledge to design a template to create instances and proliferate instances based on similarity, as a source to train data for morpheme embeddings. For evaluation, we validate the paradigmatic and syntagmatic relations for morpheme embeddings, and apply the obtained embeddings to word similarity measurement, achieving significant improvements over the classical models \footnote{The data of morpheme embeddings and word similarity measurement is available at \url{https://github.com/zi-lin/MC} for research purpose.}.

\section{Constructing Rational Knowledge of Morphemes}
In recent years, Chinese lexical and semantic ontologies such as Hantology \cite{YMC-05} and HowNet \cite{ZD-07} have exhibited valuable work related to morphemes.

One is Hantology. It's known that a Chinese character sometimes can be roughly deduced to a radical as a part of it, which may help to predict the category of the character. Some radicals themselves even stand alone, known as radical characters. Hantology exploits some 540 radical characters of \textit{ShuoWen}\footnote{\textit{Shuowen} is historically the first dictionary to analyze the structures and meanings of Chinese characters and give the rationale behind them.} as basic semantic symbols of Chinese. However, it is also limited to the use of radical characters, regardless of thousands of common Chinese characters involving about 20,000 morphemes. Hantology is obviously lacking in fine-grained representation of morphemes.

Another is HowNet. The contributors hypothesize that all the lexicalized concepts can be reduced to the relevant sememes. They, by personal introspection and examination, allegedly arrive at a set of around 2,800 language-independent sememes, ``family$|$家庭'' for example. Having no morphological analysis as the basis, the definitions of HowNet sememes are just abstract and prototypic, failing to be linked with any real and existing Chinese radical characters or characters or morphemes. Such assumptions may lead to doubt about its objectivity and coverage of Chinese semantics.

According to the above analysis, for construction of language resources, we want to ensure method objectivity as well as data coverage, and fully consider the characteristics of Chinese, i.e., the close relationship between a word and its morphemes.

For years of development, the \textit{Chinese Object-Oriented Lexicon} (\textit{COOL}) of Peking University has been in process \cite{YL-17}. We adopt the \textit{Synonymous Morpheme Set} (SMS) to denote the \textit{Morphemic Concept} (MC) and build the hierarchy of MCs. On this basis, \textit{COOL} further describes word-formation pattern and forms strict bindings between morphemes as sub-units of words and the MCs. Such rational knowledge may be applied to the fields of humanities as well as to industries. We plan to release our lexicon in the near future for research purpose.

\subsection{Constructing MCs and the Hierarchy}
\label{sec:2.1}
\subsubsection{Morpheme Extraction and Encoding}
To make sure of full coverage and fine granularity, the collection of our morphemes and words is from the 5th edition of \textit{Xiandai Hanyu Cidian }(Contemporary Chinese Dictionary, CCD)  by the Commercial Press, the most influential dictionary in China. CCD contains Chinese morphemes as well as their sense definitions, which have been carefully made by lexicographers for tens of years.

As different morphemes may originate from the identical character, we then set a unique encoding for each morpheme in CCD in the format {\tt H\_X1\_X2\_X3}, where {\tt H} represents the character as host, and {\tt X1} means that the current morpheme is the {\tt X1}$^{th}$ entry of this host in the dictionary, {\tt X2} means that there are {\tt X2} sememes in all for this entry, and {\tt X3} means that the current is the {\tt X3}$^{th}$ sememe. For example, the character ``树'' carries 4 sense definitions in the dictionary, corresponding to 4 different morphemes and sememes, and acquires encodings as follows.

\begin{table}[H]
\centering
\scalebox{0.85}{
\begin{tabular}{cl}
\hline
\textbf{Encoding} & \multicolumn{1}{c}{\textbf{Sense Definition (Sememe)}}  \\ \hline
树1\_04\_01                 & 木本植物的通称(general term of woody plant) \\
树1\_04\_02                 & 移植，栽培(to plant)                      \\
树1\_04\_03                 & 树立，建立(to set up)                     \\
树1\_04\_04                 & 姓氏(surname)                          \\ \hline
\end{tabular}}
\caption{Examples of morphemes in terms of the character ``树''}
\end{table}

We excavated data from CCD and collected a total of 8,514 Chinese characters and their 20,855 morphemes. On account of the fact that morphemes can have parts of speech (POSs), even when a morpheme is not necessarily a word, the POSs of morphemes could somehow be drawn from the POSs of words \cite{YBY-84}. We further classified all these morphemes into 13 POS types, specifically, nominal, verbal, adjectival, adverbial, numeral, classifier, pronominal, prepositional, auxiliary, conjunctional, onomatopoetic, interjection and affix morphemes. The free morphemes have postags in the dictionary, and for the bound morphemes, we manually annotated the postags of them as complement. We found that nominal, verbal and adjectival morphemes hold a total of 88.74\% of Chinese morphemes as the main body, while the rest amounts to 11.26\%.

\subsubsection{Forming MCs}

Some morphemes in Chinese actually correspond to the same or similar sememes. For example, morphemes ``树1\_04\_01'' and ``木1\_07\_01'' all refer to the meaning of \textit{general term of woody plant}. Clustering such morphemes together will help to get basic semantic units with high information density. Therefore, we are inspired to merge and represent morphemes in form of SMSs, and try to exploit such pieces of knowledge for further use.

To get reliable SMSs and considering that lexicographers used same or similar brief wordings for some sense definitions of morphemes, we measure the sememe similarity by using the co-occurrence model. The automatic clustering is just a reference for the substantial manual annotation, which ensures both the efficiency and the quality of the construction. For a particular sense definition, according to its semantic similarity score with others in descending order, and by hand-correcting, we form a corresponding SMS. Repeat this process until all the meanings of morphemes with the same POS are covered, and then turn to other POSs until all sense definitions are covered.

These SMSs just refer to the MCs of Chinese. By now, we have achieved 4,198 MCs for morphemes of the main body, including 2,018 nominal MCs, 1,630 verbal MCs and 550 adjectival MCs respectively. These MCs then form a collection of all the smallest semantic units of Chinese, showing a clear advantage of data coverage and method objectivity.

For example, here we list samples of verbal MCs in Table \ref{fig:verbal}, with the different cardinality of SMS (the number of morphemes in the SMS, \textbf{\#Mor} for short), i.e., the size of the MC. \textbf{\#MC} then refers to the total number of MCs with regards to a particular \#Mor. Here we just list one MC example for each \#Mor, along with its MC definition. For layout problem, we just remove the morpheme encodings and show characters as morphemes in the figure (so multiple occurrences of an identical character are allowed).

\begin{table}
\centering
\includegraphics[width = 0.47\textwidth]{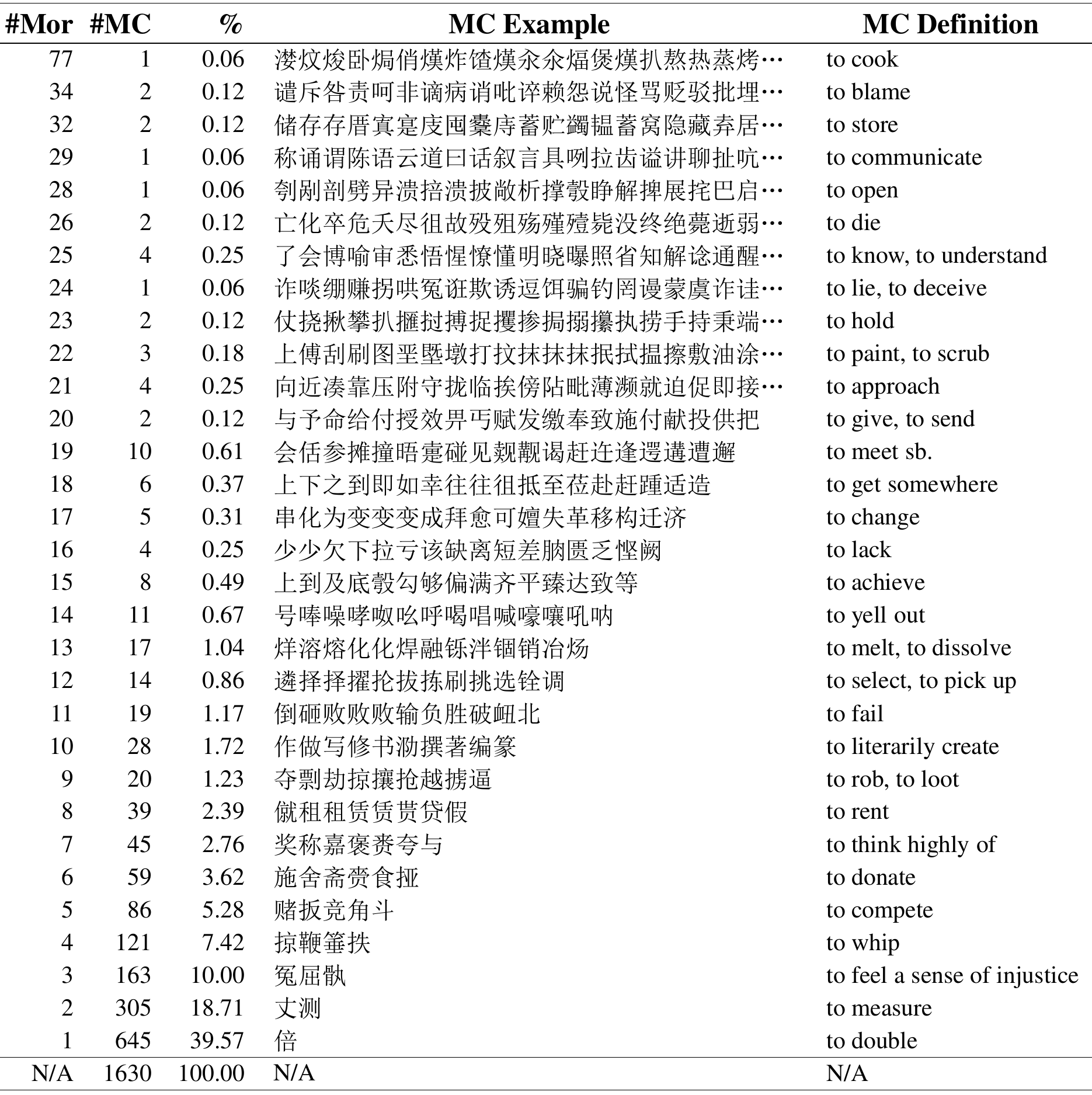}
\caption{Samples of verbal MCs and their definitions}
\label{fig:verbal}
\end{table}

\subsubsection{Building the Hierarchy of MCs}
Up to this point, the MCs are still discrete concepts. However, some MCs are highly related to others semantically. For example, as paradigmatic relation, the nominal MCs \{稗1\_02\_01, ...\} (herbage) and \{杨1\_02\_01, ...\} (xylophyta) all denote the meaning of \textit{plant}; as syntagmatic relation, the verbal MCs \{萌1\_02\_01, ...\} (to sprout) and \{茁1\_01\_01, ...\} (to grow), and the adjectival MCs \{夭2\_01\_01, ...\} (luxuriant) and \{槁1\_01\_01, ...\} (sere), are relevant to the MCs in connection with \textit{plant}. Therefore, to express those relations, a hierarchical structure for all the MCs is needed in organizing the knowledge base, with the purpose of facilitating reasoning and computing afterwards.

Inspired by WordNet \cite{miller1998wordnet}, the nominal MCs are structuralized based on hypernymy relation. As for MCs of other POSs, we are enlightened by the Generative Lexicon Theory \cite{PJ-95} to build a hierarchy where the nominal MCs are of the core structure. The verbal MCs refer to actions of the nominal, while adjectival MCs refer to attributes of the nominal. In this way, the hierarchy of the main body of Chinese morphemes, comprising the nominal, verbal and adjectival, is internally isomorphic in general.

By making use of this solution, we have obtained paradigmatic relations within the same POS and syntagmatic relations across different POSs for Chinese morphemes.

\subsection{Word-Formation Tagging}
\label{sec:2.2}
As Chinese linguists argued, morphemes as sub-units of words have particular word-formation patterns in word-building \cite{YRC-68,LS-90}. It is necessary to explore these patterns for understanding larger language units like words.

\begin{table}[H]
\centering
\scalebox{0.7}{
\begin{tabular}{llr}
\hline
\multicolumn{1}{c}{\textbf{Word-formation Pattern}} & \multicolumn{1}{c}{\textbf{Example}} & \multicolumn{1}{l}{\textbf{Percentage}} \\ \hline
定中 (Modifier-Head)&红旗 (red-flag)&37.94\%\\
联合 (Parallel)&买卖 (buy-sell)&21.90\%\\
述宾 (Verb-Object)&植树 (plant-tree)&15.62\%\\
状中 (Adverb-Verb)&广播 (widely-broadcast)&8.09\%\\
后附加 (Suffixation)&里头 (inside-$\emptyset$)&4.43\%\\
单纯词 (Noncompound)&克隆 (clone)&3.99\%\\
连谓 (Verb-Verb)&剪贴 (clip-paste)&3.28\%\\
前附加 (Prefixation)&老虎 ($\emptyset$-tiger)&1.34\%\\
述补 (Verb-Complement)&击毙 (shoot-died)&1.21\%\\
主谓 (Subject-Predicate)&地震 (earth-quake)&1.01\%\\
重叠 (Overlapping)&星星 (star-star)&0.59\%\\
介宾 (Preposition-Object)&从小 (from-youth)&0.30\%\\
名量 (Noun-Classifier)&纸张 (paper-piece)&0.15\%\\
数量 (Quantifier)&一天 (one-day)&0.11\%\\
复量 (Classifier-Classifier)&人次 (person-(per)time)&0.04\%         
\\ \hline
\end{tabular}}
\caption{Word-formation patterns and examples}
\label{tb:wf}
\end{table}

\begin{table*}
\centering
\resizebox{\textwidth}{!}{%
\begin{tabular}{ccccccc}
\hline
\textbf{Word}  & \textbf{POS} & \textbf{Word-Formation} & \textbf{$1^{st}$-MC POS} & \textbf{$2^{nd}$-MC POS} & \textbf{$1^{st}$-MC} & \textbf{$2^{nd}$-MC} \\ \hline
植树(plant-tree) & 动词(Verb) & 述宾(Verb-Object)  & 动语素(Verbal)                   & 名语素(Nominal)                   & 养1\_11\_02        & 木1\_07\_01         \\ \hline
\end{tabular}%
}
\caption{Demo of the piece of rational knowledge at morpheme level. For brevity, we use the first morpheme encoding in each MC to denote the MC itself. For example, ``养1\_11\_02" denotes the MC \{养1\_11\_02, 植1\_04\_01, ..., 莳1\_03\_01\} (to plant) where the morpheme ``植1\_04\_01'' appears within.}
\label{tb:rational knowledge of words}
\end{table*}

\begin{table*}
\centering
\scalebox{0.97}{
\begin{tabular}{|cccccccc|}
\hline
\small{\tt<B>} & \small{\tt <B>-<$1^{st}$-MC POS>} & \small{\tt <POS>} & \small{\tt <$1^{st}$-MC>} & \small{\tt <$2^{nd}$-MC>} & \small{\tt <Word-Formation>} & \small{\tt <E>-<$2^{nd}$-MC POS>} & \small{\tt <E>}\\
\small{\tt B} & \small{\tt B-Verbal} & \small{\tt Verb} & \small{\tt \{养1\_11\_02\}} & \small{\tt \{木1\_07\_01\}} & \small{\tt Verb-Object} & \small{\tt E-Nominal} & \small{\tt E}\\\hline
\end{tabular}}
\caption{The designed template (the first line) and the pseudo-sentence for the word ``植树'' (plant-tree) (the second line). {\tt <B>} and {\tt <E>} refers to the begin-position and the end-position respectively.}
\label{tb:template}
\end{table*}

For the selection of word-formation patterns, Chinese linguists generally hold two different views - one based on syntactics \cite{YRC-68} and one based on semantics \cite{LS-90}. Semantic labels have advantages of naturalness and intuition but are hard to unify and too complicated to be processed by computers. In contrast, syntactic labels are relatively simple and uniform and are somewhat consistent with syntactic structures \cite{Fu-03}. Therefore, we eventually choose the labels of word-formation geared towards syntactics to facilitate the construction of the resources. It is noteworthy that after the strict binding between morphemes and the achieved MCs, we actually acquire semantic word-formation knowledge to some extent.

After data analysis, we adopted a collection of 15 labels for Chinese word-formation tagging. The example and percentage of each word-formation pattern are listed in Table \ref{tb:wf}. By now, 52,108 Chinese disyllabic words in CCD have all been labelled with their word-formation patterns.

\subsection{Binding between Sub-units of Words and MCs}
After the work of \ref{sec:2.1} and \ref{sec:2.2}, this procedure aims to provide character-to-morpheme bindings for Chinese words, i.e., we want to assign specific MCs to morphemes as sub-units of words.

For all these 52,108 Chinese disyllabic words, we list all the possible morphemes for the first and second character, among which we choose the appropriate ones. For example, for word ``植树'' (plant-tree), the first sememe is ``栽种'' (to plant), while the second sememe is ``木本植物的通称'' (general term of the woody plant). Since each sememe corresponds to a unique morpheme encoding, the word is now formalized as \textless植1\_04\_01, 树1\_04\_01\textgreater. It is noted that each morpheme encoding belongs to a unique MC, and this actually fulfils bindings between sub-units of words and MCs. The morphemes within words would now be constrained by the hierarchy of MCs.

%Similar to word-formation tagging, characters within 52,108 disyllabic words are all mapped to unique MC. To ensure the reliability of 2.3 and 2.4, we measure the agreement among three linguistic experts, which equals 93.24\%.

Taking advantage of such lexical and semantic knowledge representation, \textit{COOL} may meet a variety of needs. In humanities, it shows potential in Chinese lexicography (e.g. concept-based browser), Chinese teaching (e.g., level evaluation standard), language study (e.g., Chinese ontology), etc. Such interdisciplinary applications can benefit from these pieces of rational knowledge \cite{YL-17}. As for NLP, to cope with the difficulties of full semantic prediction of unknown words, it can give specific lexical and semantic generation according to tasks and requirements, showing a high level of flexibility and tailorability. We leveraged the rich information of \textit{COOL} to predict word-formation patterns, morphemes and their postags within words. Our result of prediction is simple and easy for applications \cite{YT-16}.

\section{Training Distributed Representation for Morphemes}
Vector-space word representation has been successful in recent years across a variety of NLP tasks \cite{luong2013better}. Apart from rational methods in the above-mentioned application, we are motivated to implant such valuable knowledge into distributed representation. However, how to generate the so-called \textit{corpus} based on such knowledge is a central issue and makes a big challenge. And till now there has not been such practice and approach reported.

To address this issue, we design a template based on the structured rational knowledge to generate the instances, and conduct instance proliferation to exploit hierarchical information and tackle data sparseness problem. Such proliferated instances of the word by semantic word-formation, as pseudo-sentences, have thus formed a \textit{corpus} relevant to rational knowledge built in the lexicon. Then word2vec is applied to such \textit{corpus} to obtain distributed representation for morphemes.

\begin{table*}[]
\renewcommand{\arraystretch}{1}
\centering
\scalebox{0.9}{
\resizebox{\textwidth}{!}{%
\begin{tabular}{l|p{1.9cm}|p{1.9cm}||p{3.6cm}|l|}
\cline{2-5}
                          & \textbf{$C_a$}                                                         & \textbf{$C_b$}                                                         & \textbf{$\mathds{C}_a$}                                                                                                                                                          & \textbf{$\mathds{C}_a$}                                                                                                                                          \\ \hline
\multicolumn{1}{|l|}{\textbf{Seed Word}}        & 养1\_11\_02                                                             & 木1\_07\_01                                                             & \begin{tabular}[c]{@{}l@{}}养1\_11\_02 (to plant)\\ 浇1\_04\_03 (to water)\\ ... ...\\ 耕1\_02\_01 (to cultivate)\end{tabular}                                                         & \begin{tabular}[c]{@{}l@{}}木1\_07\_01 (tree)\\ 李1\_03\_01 (fruit)\\ ... ...\\ 禾1\_03\_02 (crop)\end{tabular}                                                        \\ \hline
\multicolumn{1}{|l|}{\textbf{Pseudo-Sentences}} & \multicolumn{4}{l|}{\begin{tabular}[c]{@{}l@{}}\textless养1\_11\_02, 木1\_07\_01\textgreater, \textless养1\_11\_02, 李1\_03\_01\textgreater, ..., \textless养1\_11\_02, 禾1\_03\_02\textgreater\\ \textless浇1\_04\_03, 木1\_07\_01\textgreater, \textless浇1\_04\_03, 李1\_03\_01\textgreater, ..., \textless浇1\_04\_03, 禾1\_03\_02\textgreater\\ ... ...\\ \textless耕1\_02\_01, 木1\_07\_01\textgreater, \textless耕1\_02\_01, 李1\_03\_01\textgreater, ..., \textless耕1\_02\_01, 禾1\_03\_02\textgreater\end{tabular}} \\ \hline
\end{tabular}}%
}
\caption{Demo of instance proliferation by similarity for the seed word ``植树'' (plant-tree) (\textless养1\_11\_02, 木1\_07\_01\textgreater), where $\mathds{C}_a$ and $\mathds{C}_b$ refer to a set of similar MC $C_a$ and $C_b$.}
\label{tb:proliferation}
\end{table*}

\subsection{Template Design}
Instead of using context words to predict the target word, we try to make full use of the piece of rational knowledge to generate the instantiated pseudo-sentence of morphemes. To achieve this, we propose to design a template to create the instances merely from the pieces of rational knowledge built in the lexicon. Under this assumption, word by semantic word-formation actually represents a certain and real occurrence of the combination of morphemes as in their respective MCs. Each pseudo-sentence in the so-called \textit{corpus} now refers to the instance of a word by semantic word-formation.

The piece of knowledge for use at morpheme level is shown in Table \ref{tb:rational knowledge of words}. As for such piece of information, we hence design the template as shown in the first line in Table \ref{tb:template}. Accordingly, the pseudo-sentence for the word ``植树" (plant-tree) is generated as shown in the second line.

By now, we get an instantiated pseudo-sentence of morphemes through the application of the template. Hence a total of 52,108 instances as pseudo-sentences are generated by applying the template to all the disyllabic words in the lexicon.

\subsection{Instance Proliferation}
%【按照理性知识库中现有的数据，只有五万多个，无疑是稀疏的】

To exploit hierarchical information and tackle data sparseness problem, we go on to expand our lexicon based on similarity measurement of the achieved MCs. The similarity score between two MCs ($C_1$, $C_2$) in the hierarchy of MCs is defined as

%\vspace{-1em}
\begin{equation}
sim(C_1,C_2) = \frac{2\times |path(C_1)\cap path(C_2)|}{|path(C_1)|+|path(C_2)|}
\end{equation}
%\vspace{-1em}

where $path(C)$ is the set of all the tree nodes along the path from the root to the tree node $C$.

According to the threshold, for a certain MC $C$, a set of similar MCs can be achieved as $\mathds{C}$. For every morpheme $a \in C_a$ and $b \in C_b$, if they ever happen to form a disyllabic word $ab$ in the lexicon, we then generate more pseudo-sentences between $\mathds{C}_a$ and $\mathds{C}_b$. As for the missing knowledge of POS and word-formation, it is naturally assumed to be the same as the original one. In this way, the seed word can now be proliferated into $n$ pseudo-sentences, where $n = |\mathds{C}_a \times \mathds{C}_b|$. For example, the proliferated instances we get for the seed word ``植树'' (plant-tree) (\textless养1\_11\_02, 木1\_07\_01\textgreater) are listed in Table \ref{tb:proliferation}.

We set the threshold equal to $0.85$ in the experiments and finally get a total of \textbf{54,880,628} pseudo-sentences from \textbf{52,108} real disyllabic words as the seeds input.

\subsection{Data Training}
Word2vec \cite{mikolov2013efficient} is an algorithm to learn distributed word representation using a neural language model. It has two models, the continuous bag-of-words one (CBOW) and the skip-gram one. 

In this paper, we train the distributed representation for morphemes based on CBOW, which aims at predicting the target word given context words in a sliding window. For morpheme embeddings on these 54,880,628 pseudo-sentences, we set the dimension to 20 and context window size to 3 to include all the rational knowledge when the MC is the target word.

\section{Experimental Results and Evaluation}
By the above approach we proposed, the rational knowledge of morphemes is now implanted into distributed representation. To evaluate such representation, intrinsic evaluation, such as paradigmatic and syntagmatic relations among morphemes, and extrinsic evaluation like word similarity measurement are taken into consideration.

%【在这个例子之后做一个数据的对比，特别强调我们还有一种验证】
\begin{figure}[H]
\centering
\begin{minipage}[t]{0.48\textwidth}
\centering
\adjincludegraphics[width=2.3in, angle = 270, trim=2.1cm 0.6cm 0cm 1.2cm, clip]{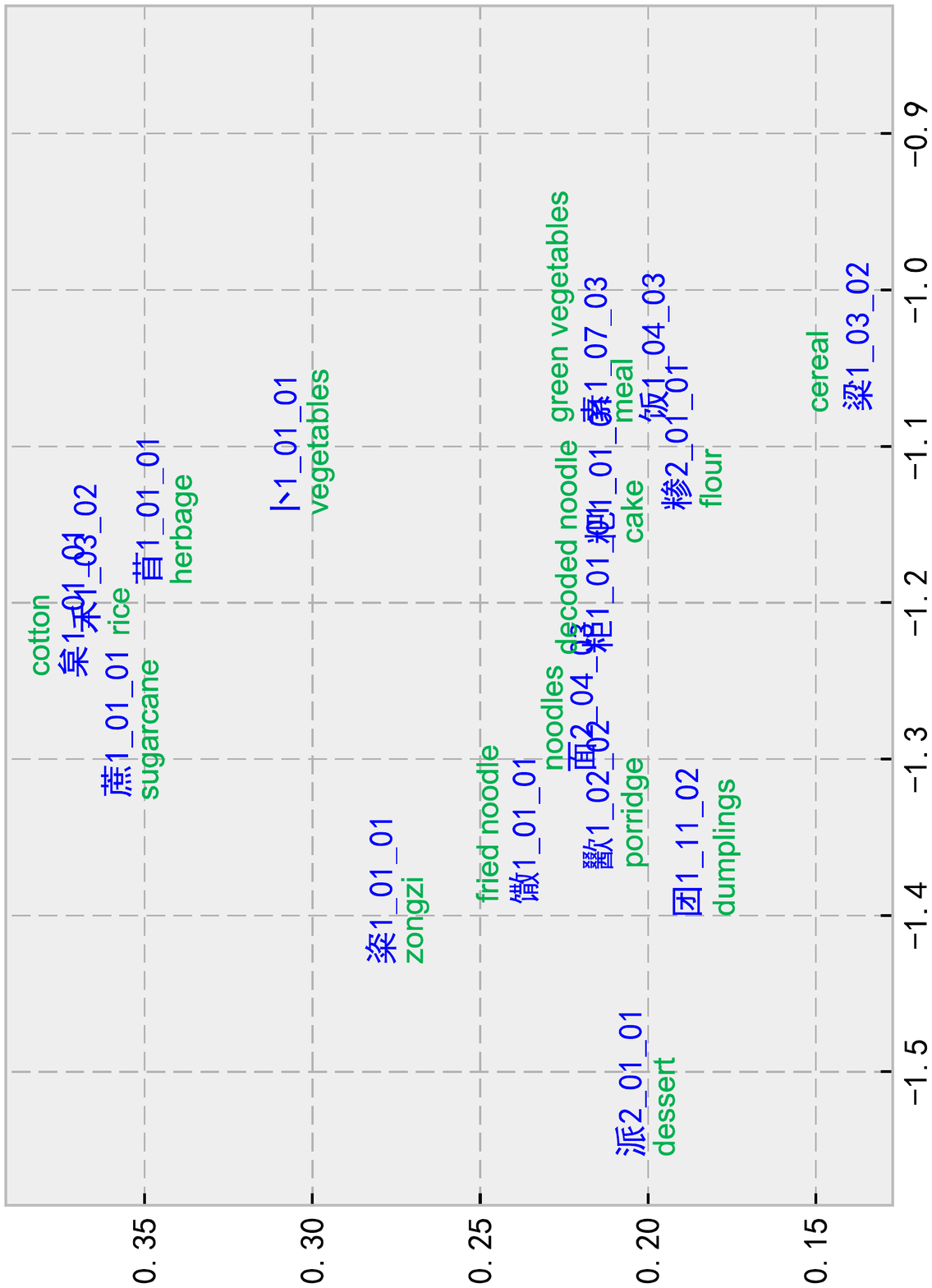}
\end{minipage}\\
\begin{minipage}[t]{0.48\textwidth}
\centering
\adjincludegraphics[width=2.3in, angle = 270, trim=2.1cm 0.6cm 0cm 1.2cm, clip]{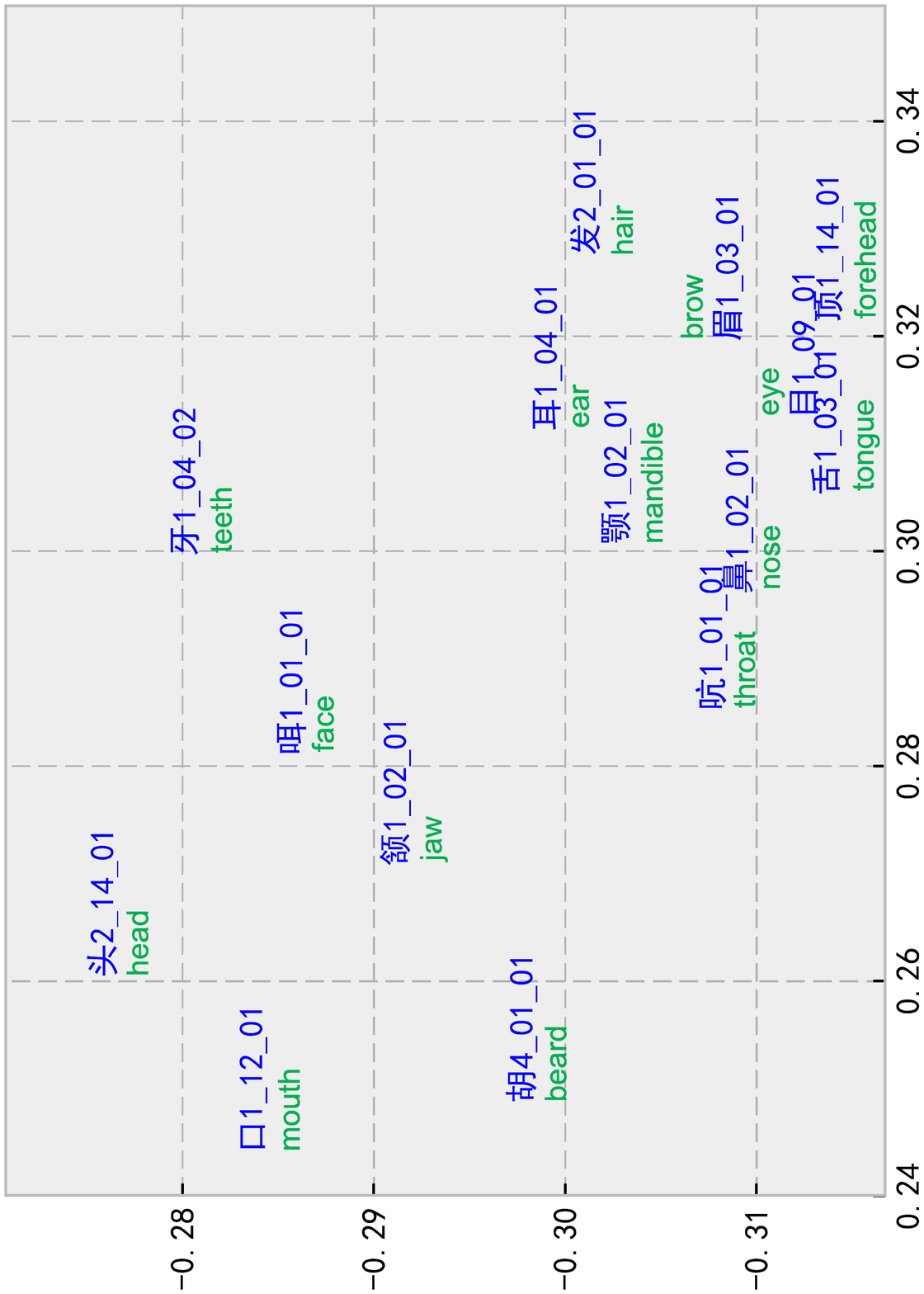}
\end{minipage}
\caption{Illustration of paradigmatic MCs in two-dimension planes}
\label{gif: PCA}
\end{figure}

\subsection{Paradigmatic Relation Validation}
% \usepackage{graphicx}
\begin{table*}[]
\centering
\scalebox{1}{
\resizebox{\textwidth}{!}{%
\begin{tabular}{lll}
\hline
\multicolumn{1}{c}{\textbf{Morpheme}}                            & \multicolumn{1}{c}{\textbf{MC}}  & \multicolumn{1}{c}{\textbf{Nearest MCs}}                                                                                                                                                                              \\ \hline
法1\_07\_01 (law)                                                  & \{宪1\_03\_02, 制1\_05\_04, ..., 辟4\_01\_01\} & 
\begin{tabular}[c]{@{}l@{}}\{命2\_03\_01, 号3\_04\_01, ..., 禁1\_04\_03\} (command)\\ \{禅1\_02\_02, 佛3\_05\_03, ..., 藏1\_02\_02\} (Dharma)\\ \{墨1\_10\_08, 黥1\_02\_02, ..., 刑1\_03\_02\} (punishment)
\end{tabular}                    \\ \hline
\begin{tabular}[c]{@{}l@{}}法1\_07\_02 (method)\end{tabular}     & \{道1\_12\_03, 方3\_02\_01, ..., 筹1\_03\_03\} & 
\begin{tabular}[c]{@{}l@{}}\{典1\_06\_01, 度2\_15\_06, ..., 师1\_07\_02\} (model)\\ \{揆1\_04\_02, 理1\_07\_02, ..., 谛1\_02\_02\} (theory and principle)\\ \{准2\_06\_01, 标1\_10\_04, 杠1\_07\_07\} (standard)\end{tabular}            \\ \hline
\begin{tabular}[c]{@{}l@{}}法1\_07\_04 (to emulate)\end{tabular} & \{则1\_05\_03, 宗1\_08\_05, ..., 象2\_02\_02\} & 
\begin{tabular}[c]{@{}l@{}}\{问1\_06\_01, 叩1\_04\_03, ..., 难1\_02\_02\} (to address inquiries)\\ \{绎1\_01\_01, 验1\_03\_01, ..., 搜1\_02\_02\} (to search)\\ \{造1\_03\_02, 编1\_09\_05, ..., 蔑2\_01\_01\} (to make up)\end{tabular} \\ \hline
\end{tabular}%
}}
\caption{Different morphemes with identical character ``法'' and their nearest MCs}
\label{tb: fa}
\end{table*}

\begin{table*}[]
\centering
\scalebox{0.8}{
\resizebox{\textwidth}{!}{%
\begin{tabular}{ll}
\hline
\multicolumn{1}{c}{\textbf{MC}}          & \multicolumn{1}{c}{\textbf{Top-Related MCs}}                                                                                                                                                      \\ \hline
\{骏1\_01\_01, 马1\_03\_01, ..., 驹1\_02\_02\} (horse)  & \begin{tabular}[c]{@{}l@{}}\{骏1\_01\_01, 马1\_03\_01, ..., 驹1\_02\_02\} (horse)\\ \{镫1\_01\_01, 鞍1\_01\_01, ..., 鞒1\_01\_01\} (saddle)\\ \{兵1\_05\_02, 军1\_03\_01, ..., 卒1\_03\_01\} (soldier)\end{tabular} \\ \hline
\{鸡1\_02\_01, 鸭1\_01\_01,..., 鹅1\_01\_01\} (fowl)   & \begin{tabular}[c]{@{}l@{}}\{仔3\_01\_01, 子1\_13\_08, ..., 雏1\_02\_02\} (chick)\\ \{野1\_07\_04\} (wild)\\ \{坤1\_02\_02, 母1\_06\_03, ..., 牝1\_01\_01\} (female)\end{tabular}                                 \\ \hline
\{牛1\_04\_01, 牦1\_01\_01, ..., 犊1\_01\_01\} (cattle) & \begin{tabular}[c]{@{}l@{}}\{乳1\_05\_03, 奶1\_03\_02\} (milk)\\ \{牛1\_04\_01, 牦1\_01\_01, ..., 犊1\_01\_01\} (cattle)\\ \{土1\_07\_01,垆1\_01\_01, ..., 壤1\_03\_01\} (soil)\end{tabular}                       \\ \hline
\end{tabular}%
}
}
\caption{Different MCs and their related MCs in word-building}
\label{tb:syntagtic relations}
\end{table*}

To evaluate the effectiveness of the new approach, we use Principal Component Analysis (PCA) to conduct dimensional reduction on morpheme embeddings to show paradigmatic relations as internal knowledge input. The results are illustrated in Figure \ref{gif: PCA}.

We further take different morphemes with the identical character for example and list their corresponding MCs. Table \ref{tb: fa} illustrates the nearest 3 MCs of each morpheme for observation. It can be observed that the MCs with similar meanings are naturally gathered together in general.

\subsection{Syntagmatic Relation Validation}
In addition to paradigmatic relations, we also explore the syntagmatic relations as internal knowledge input, i.e., which morphemes are more likely to form words. 

For each MC, we predict its context words, which are the rational knowledge already input, and extract the most probable MCs in the context words.

We list some of the MCs and 3 of their top-related MCs in Table \ref{tb:syntagtic relations}, from which it can be observed that given a specific MC, which MCs are prone to be involved in word-building. From the table, we notice that the cases of combination obtained from morpheme embeddings are quite consistent with human judgment. Take MC \{骏1\_01\_01, 马1\_03\_01, ..., 驹1\_02\_02\} (horse) for example, many words can be formed between the MC and its top-related MCs, such as ``骏马'' (steed), ``马驹'' (foal), ``马鞍'' (saddle) and ``骑兵'' (cavalryman).

\subsection{Word Similarity Measurement}
The above are all intrinsic evaluation. For extrinsic evaluation, word similarity computation is an appropriate task. In order to compute the semantic distance between word pairs, many previous works take words as basic units and learn word embeddings according to external knowledge (contexts), ignoring the internal knowledge of words (morphemes and word-formations). However, as we argued, the internal knowledge also plays an important role in Chinese. The ideal way to measure semantic similarity may be to combine external and internal knowledge together.

\begin{table}[H]
\centering
\scalebox{1}{
\begin{tabular}{lll}
\hline
\textbf{Word-Formation Pattern} & \textbf{$1^{st}$-MC} & \textbf{$2^{nd}$-MC} \\ \hline
后附加 (Suffixation)&1&0\\
述补 (Vreb-Compliment)&0.8&0.2\\
述宾 (Verb-Object)&0.6&0.4\\
联合 (Parallel)&0.5&0.5\\
单纯词 (Noncompound)&0.5&0.5\\
定中 (Modifier-Head)&0.45&0.55\\
状中 (Adverb-Verb)&0.45&0.55\\
主谓 (Subject-Predicate)&0.4&0.6\\
前附加 (Prefixation)&0&1\\
\hline
\end{tabular}}
\caption{Weight assignments for different word-formation patterns}
\label{tb:weight}
\end{table}

As the meaning of a word is contributed by the morphemes as its sub-units, which now refers to the MCs in the hierarchy, we assign different weights to the morphemes appearing in different word-formation patterns. For example, for ``定中'' (Modifier Head) structure, the head will contribute more to the meaning of the word, while for ``前附加'' (Prefixation) structure, the prefix can hardly be related to the meaning of the word. Eventually, 9 types of word-formation pattern in the test sets (see description below) are assigned with different weights for the morphemes, as shown in Table \ref{tb:weight}.

Based on this, we try to obtain word embedding for each word by a weighted average of its $n$ morpheme embeddings. It is calculated as $v_i = \sum_{k=1}^{n} w_{i_k} \times c_{i_k}$, where $v_i$ stands for the word vector of the $i^{th}$ word in the lexicon, $w_{i_k}$ is the weight assigned by the above table and $c_{i_k}$ is the morpheme embedding which the $k^{th}$ character corresponds to. This is how our MC model will work on word similarity by purely exploiting internal knowledge.

As for CBOW and skip-gram, we use the corpus of \textit{Baidu Encyclopedia}, which contains 203.69 million words in total. In the experiments, the dimension is set to 50, and the context window size is set to 5. Cosine similarity is applied to measure word similarity score for models of CBOW, skip-gram and MC individually. We also combine similarity scores obtained from the classical model (CBOW and skip-gram respectively) and MC model with the same weight assignment, namely the hybrid models of CBOW+MC and skip-gram+MC.

In the experiments, wordsim-296 \cite{jin2012semeval} and PKU-500 \cite{wu2016overview} are used as evaluation datasets. We extract the disyllabic words in the datasets and get a total of 141 word pairs in wordsim-296 and 232 word pairs in PKU-500 respectively. These serve as the test sets for word similarity measurement. Spearman's correlation $\rho$ \cite{myers2010research} is then adopted to evaluate all the outputs on the test sets. The experimental results of the 5 models are shown in Table \ref{tb:performance}.

\begin{table}[H]
    \centering
	\scalebox{1}{
	\begin{tabular}{rcc}
	\hline
	\multicolumn{1}{c}{\textbf{Model}} & \textbf{wordsim-296} & \textbf{PKU-500}          \\ \hline
	CBOW & 57.43& 34.82            \\
	Skip-gram & 62.17& 40.19            \\ \hline
	%CWE & 58.60 & 39.25            \\ 
    %SE-WRL & 61.40 & 40.89               \\\hline
	MC & 46.28  & 30.57            \\ 
	CBOW+MC   & 64.35  & 42.74            \\
	Skip-gram+MC    & \textbf{67.58} & \textbf{45.91} \\\hline
	\end{tabular}}
	\caption{Evaluation results on wordsim-296 and PKU-500 ($\rho \times 100$)}
	\label{tb:performance}
\end{table}

%【MC跟两种经典方法的对比，强调完全不用语料就可以达到，这种做法此前未有报道】

%【加上清华的数据的引用对比和说明，强调in跟ex的结合显示了额外的performance，这还是比较鼓舞的】

Note that our morpheme embeddings are trained with only 52,108 original pieces of semantic word-formation knowledge (approximately 2.79 MB of storage as in the experiments), without a corpus of data harnessed as before. The MC model, by purely exploiting such internal knowledge, alone achieves a fairly good performance, compared with the classical models. Furthermore, experiments on test sets show that the hybrid models of CBOW+MC and Skip-gram+MC, by exploiting external and internal knowledge, achieve significant improvements over the classical models by more than 5 Spearman scores or 8 percentage points. This indicates that both sources of knowledge are very valuable and highly complementary in expressing the meanings of words.

\begin{figure}[]
  \centering 
    \includegraphics[width = 3.3in]{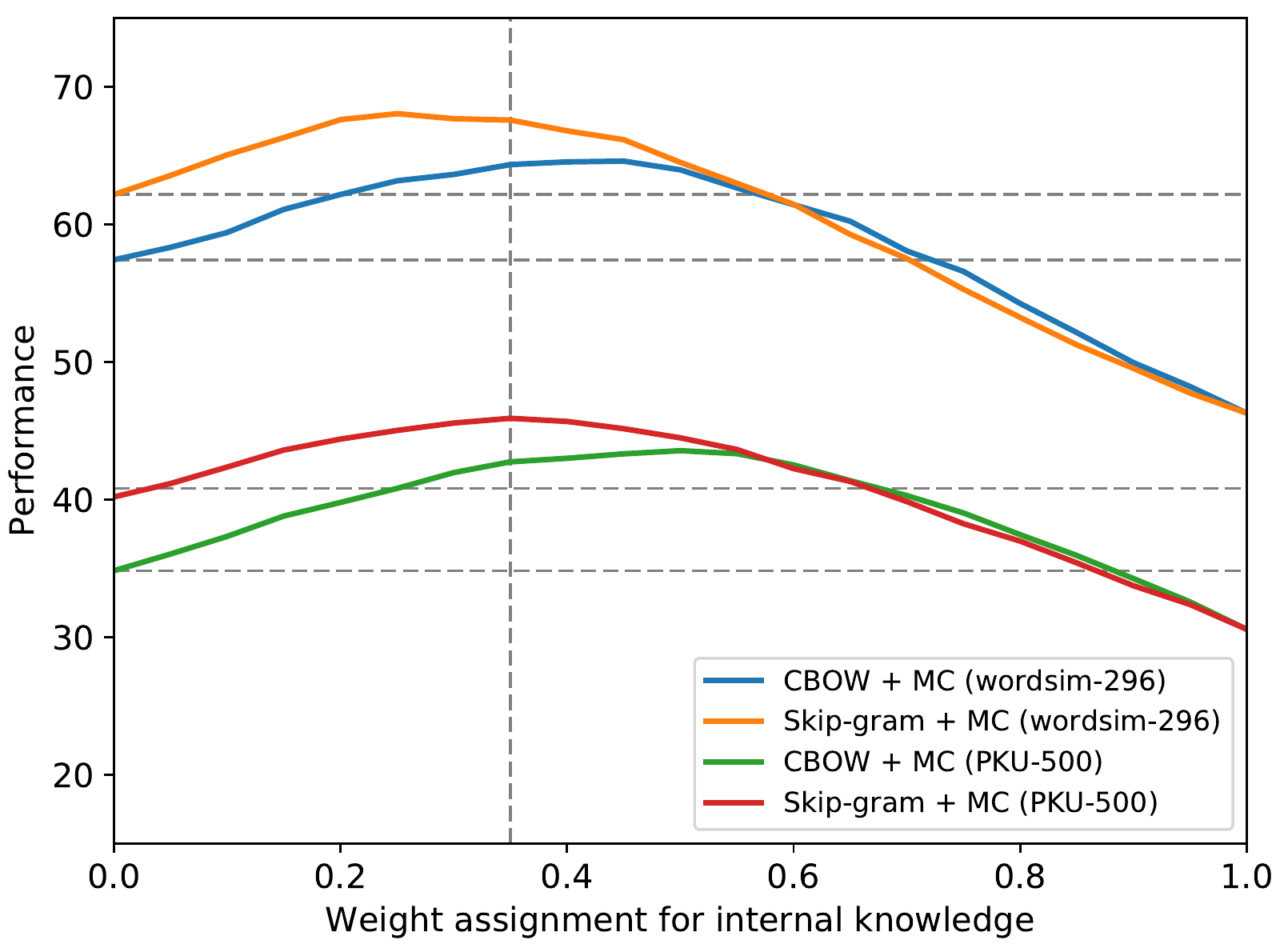}
	  \caption{Correlation between weight assignment for internal knowledge and performance}
    \label{fig:weight and performance}
\end{figure}

As for the combination for hybrid models, we set different weight assignments for similarity scores obtained from the classical model (CBOW and skip-gram respectively) and MC model to explore the cases. The correlation between internal knowledge adopted and performance is shown in Figure \ref{fig:weight and performance}. Considering the global optimum for both test sets, the most ideal weight assignment for internal and external knowledge is $0.35:0.65$ in the experiments, which has been adopted and yielded the results in Table \ref{tb:performance}.

\section{Conclusion}
In this paper, after constructing the Chinese lexical and semantic ontology based on word-formation, we try to implant the structured rational knowledge into distributed representation at morpheme level without using any text corpus. For evaluation, we validate the paradigmatic and syntagmatic relations of morpheme embeddings, and apply the obtained embeddings to word similarity measurement, achieving significant improvements over the classical models by more than 5 Spearman scores or 8 percentage points.

The key contributions of this work are as follows: (1) We, for the first time, put forward an approach to implanting the structured rational knowledge into distributed representation by merely using the lexicon. As the form of such piece of knowledge is common to most knowledge bases, we actually present an inspiring way of obtaining distributed representation for the desired language units described in the lexicons. (2) For parataxis languages like Chinese, morphemes as the basic units play an important role in expressing the exact meanings of words. It is a convenient way by obtaining unambiguous morpheme embeddings simply based on the descriptions in the lexicon, which naturally avoids heavy disambiguation in the corpus as before \cite{P18-1230,DBLP:conf/emnlp/LuoLHXSC18}.

Currently, we focus on the original meanings of Chinese disyllabic words, which make up the majority of the vocabulary of CCD. However, some words may have metaphoric or transferred meanings, or comprise of more than two characters. Such work is in progress in our group according to the solution. Also, to gain better word embeddings for certain tasks, the topic of compositionality of word embeddings is reserved for further research. After completion of these works, we hope to release the \textit{COOL} system.

\section{Acknowledgments}
This work is supported by National Basic Research Program of China (973 Program, No. 2014CB340504) and National Social Science Fund of China (Key Project, No. 12\&ZD119; General Project, No. 16YY137). We thank the anonymous reviewers for their helpful comments. The corresponding author of this paper is Yang Liu.
\end{CJK}
\bibliographystyle{aaai}
\bibliography{aaai19}
\end{document}